\documentclass[5p,times]{elsarticle}
\usepackage{todonotes}
\usepackage{amsmath}
\usepackage[hidelinks]{hyperref}

\usepackage{multirow}
\usepackage{algpseudocode}
\usepackage{algorithm}
\algdef{SE}[DOWHILE]{Do}{doWhile}{\algorithmicdo}[1]{\algorithmicwhile\ #1}%

\usepackage{amssymb}
\usepackage[T1]{fontenc}
\usepackage[utf8]{inputenc}

\usepackage{colortbl}

\journal{Future Generation Computer Systems. Accepted version: \url{https://doi.org/10.1016/j.future.2023.11.003}}

\begin{document}

\begin{frontmatter}

\title{ColabNAS: Obtaining lightweight task-specific convolutional neural networks following Occam's razor}

\author[inst1]{Andrea Mattia Garavagno}\ead{AndreaMattia.Garavagno@santannapisa.it (corresponding author)}
\author[inst1]{Daniele Leonardis} 
\author[inst1]{Antonio Frisoli} 

\affiliation[inst1]{organization={Institute of Mechanical Intelligence, Scuola Superiore 
Sant’Anna of Pisa},%Department and Organization
            addressline={Piazza Martiri della Libertà, 33}, 
            city={Pisa},
            postcode={56127}, 
            state={Tuscany},
            country={Italy}}

\begin{abstract}
The current trend of applying transfer learning from convolutional neural networks (CNNs) trained on large datasets can be an overkill when the target application is a custom and delimited problem, with enough data to train a network from scratch. On the other hand, the training of custom and lighter CNNs requires expertise, in the from-scratch case, and or high-end resources, as in the case of hardware-aware neural architecture search (HW NAS), limiting access to the technology by non-habitual NN developers.

For this reason, we present ColabNAS, an affordable HW NAS technique for producing lightweight task-specific CNNs. Its novel derivative-free search strategy, inspired by Occam's razor, allows to obtain state-of-the-art results on the Visual Wake Word dataset, a standard TinyML benchmark, in just 3.1 GPU hours using free online GPU services such as Google Colaboratory and Kaggle Kernel.
\end{abstract}

\begin{highlights}
\item Hardware-aware neural architecture search algorithm for task-specific convolutional neural networks
\item A novel low-cost derivative-free search strategy inspired by Occam's razor
\item State-of-the-art results on the Visual Wake Word dataset in just 3.1 GPU hours
\item Able to be executed on free subscription online GPU services
\item Convolutional Neural Networks for low-RAM microcontrollers (e.g. 40 kiB)
\end{highlights}


\end{frontmatter}


\section{Introduction}
\label{sec:introduction}

Task-specific convolutional neural networks (CNNs) are enabling the rise of next-generation smart wearable systems and distributed sensors. The limited size of the problem to be solved allows for lightweight models. However, developing from scratch a good lightweight neural network model is not easy. A recent research area named hardware-aware Neural Architecture Search (HW NAS) is facing the problem, of tailoring the search to the precise resources of the target hardware. As of today, open-source projects such as MCUNet \cite{mcunet} and Micronets \cite{micronets} are able to produce state-of-the-art TinyML models in around 300 GPU hours. Nevertheless, lightweight task-specific CNN are commonly designed using Transfer Learning (TL) \cite{DeepTransferLearning}.

For example \citet{Chest_X_ray}, in 2022, proposed a lightweight task-specific CNN able to diagnose COVID-19 by evaluating chest X-rays. To design the network the TL procedure is applied. MobileNetV2 \cite{Mobilenetv2} is used as the model backbone. 

A similar approach is applied by \citet{remote_ship_detection}, in 2022, to detect ships in spaceborne synthetic aperture radar (SAR) images. In this case, MobileNetV2 is used as a backbone for a YOLOv4-LITE model. 

In the same year, TL is also applied by \citet{arc_detection} to develop a model based on Efficientnet-B1 \citep{EfficientNet}, able to detect arc fault in photovoltaic systems, by analyzing power spectrum images; by \citet{wheat_detection_and_count} to develop a model based on MobileNetV2 able to detect and count wheat heads by analyzing pictures; and by \citet{oral_cancer_detection} to develop a model based on EfficientNet-B0 \citep{EfficientNet} able to early detect oral cancer by pictures of the oral cavity.

Ragusa et al. used MobileNets as a backbone for performing image polarity detection using visual attention \cite{polarity_detection} and for performing affordance detection \cite{affordance_detection} on embedded devices.

The above represents a brief overview of the rich literature involving lightweight task-specific neural network designs, using TL, presented just in 2022.

TL uses the knowledge acquired by a CNN in solving a problem, to solve another similar one, reducing the training data and the required time. This technique helps when there is not enough data and/or time to design a model from scratch. On the contrary, TL can be an overkill: task-specific problems can be coped with by significantly lighter CNNs. Even the lightest models, used in the TL procedure, are trained on large-scale datasets (i.e. ImageNet1k by \citet{ImageNet}). In addition, in the case of application to specific problems, related datasets are often already available for the end-users. Generating a CNN from scratch seems then a convenient choice. Yet the search cost of HW NAS, or the cost of manual design, is still too high compared to the time required by TL, possibly explaining the prominent use of TL over HW NAS in the literature. 

In this paper, we propose ColabNAS, an affordable HW NAS technique for producing lightweight task-specific CNNs. It uses a novel derivative-free search strategy, inspired by Occam's razor, that allows it to obtain state-of-the-art results on the Visual Wake Word dataset \cite{vww} in just 3.1 GPU hours using free online GPU services such as Google Colaboratory and Kaggle Kernel. Such a feature inspired its name, which wants to emphasise the ability to be run on free subscription online services, making its use available to everyone. We believe this feature is critical to foster the use of CNNs in the variety of application tasks they can adapt, especially in the field of embedded wearable and distributed devices, and implemented by a heterogeneous population of end-users and researchers. 

In this paper, we present details of the proposed technique and experimental validation on five different classification problems on online-available datasets. Then, the results obtained on the Visual Wake Word dataset are compared with state-of-the-art HW NAS methods. We provide online access to experimental data and code in the form of Google Colaboratory notebooks at this \href{https://drive.google.com/drive/folders/14wkOeM7TcNkZLpWwrVJRjHrxt0LG_7Ad?usp=sharing}{\color{blue}{link}}. Code is also available on GitHub at the following \href{https://github.com/AndreaMattiaGaravagno/ColabNAS}{\color{blue}{link}}.

\section{Related Works}
\label{sec:related_works}
In recent years standard benchmarks for tiny machine learning applications were established \cite{TinyML_benchmarks}. Imagenet1k, by \citet{ImageNet}, and the Visual Wake Words, by \citet{vww}, datasets were chosen to measure the performance of the most recent efforts of trying to implement deep-learning techniques for computer vision tasks on commodity hardware. 

These efforts followed the path opened by manually designed lightweight models such as MobileNets \cite{Mobilenet} \cite{Mobilenetv2} \cite{Mobilenetv3}, SqueezeNet \cite{Squeezenet}, ShuffleNets \cite{Shufflenet} \cite{Shufflenetv2}, and tried to automatize the design using NAS techniques, taking into consideration hardware constraints such as Flash and SRAM occupancy, or latency, giving birth to HW NAS.

Several HW NAS make use of reinforcement learning to search for the optimal architecture under the constraints set by the target hardware. A controller produces sample architectures and is rewarded based on the validation accuracy and the hardware cost. This is the case of MNASNet \cite{MNASnet}, FPNet \cite{FPNet}, Codesign-NAS \cite{CodesignNAS}, and \cite{ASIC}. This approach commonly implies a training phase for each architecture generated, which leads to high search costs. In particular, MNASNet \cite{MNASnet} tries to find the Pareto optimal solution of an objective function penalized by hardware constraints using reinforcement learning. The search space is hierarchical; at each iteration, an RNN, the controller, produces a sample architecture which is trained for evaluating the accuracy and then executed on the target hardware to evaluate the latency. MNASNET declares a search cost of 40,000 GPU hours.

Such issue has been addressed in the literature by proposing an over-parametrized network, known as supernetwork; it contains all the models belonging to the search space as sub-networks, which share weights. This solution requires only the training of the supernetwork, thus reducing the search time. Then, techniques like evolutionary algorithms \cite{mcunet} \cite{ONCEFORALL} \cite{NASCaps} and gradient descent methods \cite{Fbnet} \cite{DARTS} \cite{micronets} \cite{DDPNAS} \cite{DLWNAS} \cite{Proxylessnas} are commonly used to search for the best sub-network. In details, MCUNet is composed of two parts an HW-NAS technique called TinyNAS, and a framework for executing deep learning on microcontrollers called TinyEngine. The search space is built considering the hardware constraints. Flash, SRAM occupancy and latency are taken into account and measured directly on the target hardware. Instead Micronets, also targeting microcontrollers, avoids the hardware in the loop by computing RAM and Flash memory occupancy and using op count as a viable proxy to latency.

Even if the search time has been drastically reduced, a supernetwork still requires a large amount of resources to be trained. ColabNAS introduces a novel approach to HW NAS that limits search time and resource usage. The low resource usage is guaranteed by the proposed search space, which contains regular architectures with small footprints that require few resources to be trained. The low search cost is achieved by adopting the novel derivative-free search strategy proposed in this paper, which reduces the number of solutions explored by leveraging the Occam's razor.

\section{ColabNAS}
\label{sec:technique}
This section describes the developed technique to search lightweight, task-specific CNNs.

%Every HW NAS technique has three unique identification factors: the search space, the corresponding optimization problem formulation and the solution search strategy. 

Every HW NAS technique has three unique identification factors: the search space, the optimization problem formulation and the search strategy.

The search space defines the set of possible solutions which the optimization problem may have. The optimization problem establishes the boundaries of the search and the logic used to evaluate possible solutions. Finally, the search strategy describes the way the search space is explored, i.e. how the problem solution is found.

\subsection{Search Space and Problem Formulation}
\begin{figure}[ht]
    \centering
    \includegraphics[width=0.4\textwidth]{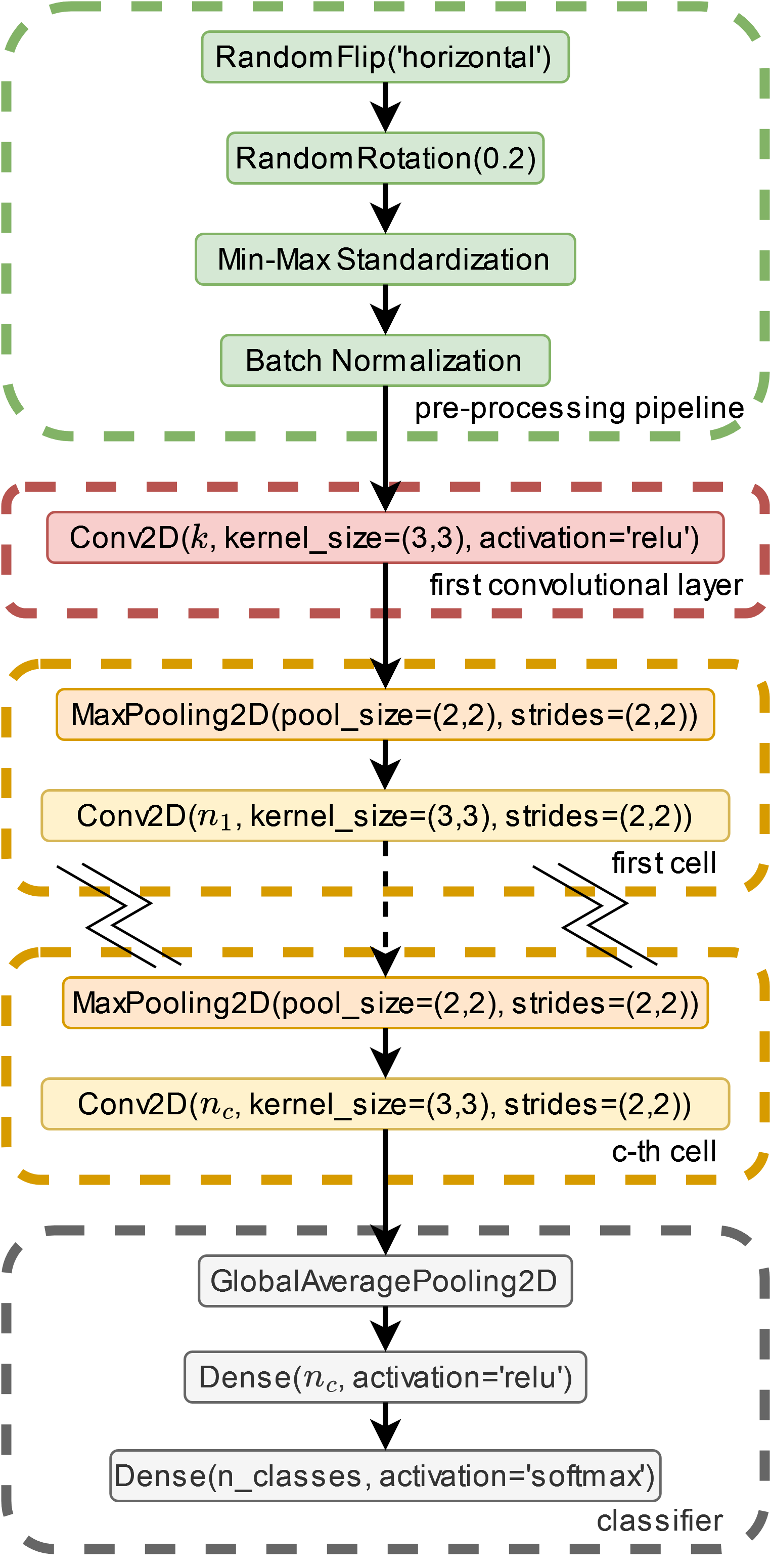}
    \caption{Detailed graphical representation of the generic network architecture. $n_{c}$ represents the number of kernels used in the c-th cell. Instead, the dotted connection represents a generic number of cells added. The deep dense layer has a number of neurons equal to the kernels used in the last cell added.}
    \label{fig:generic_architecture}
\end{figure}

There are three types of search spaces: layer-wise, cell-wise and hierarchical. ColabNAS uses a cell-wise search space. It starts from a single bidimensional convolutional layer. Then, it continues by stacking couples of pooling and convolutional layers, like in the VGG16 architecture \cite{VGG16}, which form a single cell as shown in figure \ref{fig:generic_architecture}, until the network's generalization capability increases. 

The stopping criterion takes inspiration from the problem-solving principle that ``entities should not be multiplied beyond necessity'', also known as Occam's razor, attributed to English Franciscan friar William of Ockham, who lived between the thirteen and the fourteenth century.

The number of kernels of each cell added is determined by equation \ref{eq:kernels}, where $k$ represents the number of kernels used in the first convolutional layer. It is inspired by the procedure used in the VGG16 architecture \cite{VGG16}, where a cell doubles the number of kernels with respect to the previous cell. However, in the proposed approach this amplification is modulated cell after cell, to limit the parameters’ growth.

\begin{equation} \label{eq:kernels}
\begin{aligned}
n_{c} = \begin{cases}
   \multicolumn{1}{@{}c@{\quad}}{k} & if \quad c = 0\\
   \left \lceil{ (2 - \sum_{i=1} ^{c - 1} 2^{-i}) \cdot n_{c - 1} }\right \rceil & if \quad c \geq 1
\end{cases}
\end{aligned}
\end{equation}  

The search space is constrained by the network's peak RAM occupancy, Flash occupancy and by the number of multiply and accumulate operations (MACC) which is used as a rough estimate of latency as suggested by \citet{micronets}. 

\begin{equation} \label{eq:problem}
\begin{aligned}
P1: \begin{cases}
\max f(x)\\
   \phi_{R}(x) \leq \xi_{R} \\
   \phi_{F}(x) \leq \xi_{F}    \\
   \phi_{M}(x) \leq \xi_{M} \\
   \xi_{R}, \xi_{F}, \xi_{M} > 0 
\end{cases}
\end{aligned}
\end{equation}

This leads to the problem formulation presented in equation (\ref{eq:problem}), where function $f$ returns the maximum validation accuracy obtained during training; function $\phi_{R}$ the network's peak RAM occupancy; function $\phi_{F}$ the network's Flash occupancy; function $\phi_{M}$ the network's number of MACC. These depend on the number of kernels used in the first layer $k$, the number of cells added $c$, and the magnitude of the network's input size $s$. The latter is omitted in the formulation since it is considered fixed during the search, hence the search variable $x = (k, c)$ is defined. $f(x)$ values come from training. The feasible search space will be referred as $\Omega$. 

\subsection{Network's Architecture details}
Inspired by VGG16 \cite{VGG16} we decided to adopt convolutional layers with 3x3 kernels and zero padding to preserve the input size; pooling layers with 2x2 receptive field and (2, 2) stride. 

The convolutional base output is reduced by applying the 2D Global Average Pooling operator to improve the model's generalization capability \cite{Lin2014NetworkIN}. Subsequently, a deep fully connected layer, having the number of neurons equal to the number of kernels of the last convolutional layer, further elaborates the reduced features. Finally, a single fully connected layer classifies the extracted features. 

A pre-processing pipeline is included in the network's architecture. At first it applies min-max standardization, to improve gradient descent convergence rate \cite{SHANKER1996385}. Then, it uses batch normalization to stabilize and speed up the training \cite{ioffe2015batch}. Data augmentation is also applied. Both horizontal flips and random rotations are used. Figure \ref{fig:generic_architecture} graphically represents the generic architecture used as a search space.

\subsection{Search Strategy}

The search strategy explores the search space in two steps. First, as stated before, given a starting number of kernels of the first layer, it starts to add cells until the network's generalization capability increases, according to Occam's razor, or until the network respects hardware constraints. Then it repeats the latter process changing the number of kernels of the first layer according to equation \ref{eq:number_of_kernel_of_the_first_layer}. 

\begin{equation}\label{eq:number_of_kernel_of_the_first_layer}
\begin{aligned}
k_{j} = \begin{cases}
 k_{0} & if \quad j = 0 \\
 2 \cdot k_{0}    & if \quad j = 1 \\
 2 \cdot k_{j - 1}    & if \quad f(k_{1}^{*}, c_{1}^{*}) > f(k_{0}^{*}, c_{0}^{*})    \\
 \frac{1}{2} \cdot k_{j - 1}  & if \quad f(k_{1}^{*}, c_{1}^{*}) \leq f(k_{0}^{*}, c_{0}^{*})    
 \end{cases}
\end{aligned}
\end{equation}

If the hardware selected for deployment allows, the process is repeated with double the number of kernels. If the solution found at the second iteration $(k_{1}^{*},c_{1}^{*})$ is better than the previous one $(k_{0}^{*},c_{0}^{*})$, the process continues in the same way until performance improves, if the hardware limits are not reached. Otherwise, the process continues by halving down the initial number of kernels until performance degradation is met. Also this time Occam's razor is respected. Entities are not multiplied beyond necessity.

Such a search strategy can be interpreted as a custom derivative-free method for solving the constrained optimization problem presented in equation \ref{eq:problem}. Algorithm \ref{alg:search} describes such an interpretation. 

\begin{algorithm}[!h]
\caption{Search Strategy}\label{alg:search}
\begin{algorithmic}
\Require $x_{0} \in \Omega$
\Ensure $\begin{cases}
   f(\widetilde{x_{j}}) > f(\widetilde{x_{j - 1}}), \widetilde{x_{j}} \in \Omega & \text{always} \\
   f(\widetilde{x_{j + 1}}) \leq f(\widetilde{x_{j}}) & \text{if } \widetilde{x_{j + 1}} \in \Omega
\end{cases}$
\State $\epsilon \gets 0.005$, $k_{0} \gets 4$, $x_{0} \gets (k_{0}, 0)$, $d \gets (1, 0)$, $t_{0} \gets x_{0}$, $j \gets 0$
\State $\widetilde{x_{0}}$ = \text{\scriptsize EXPLORE\_NUM\_CELLS}($x_{0}$)
\State $x_{1} = x_{0} + t_{0} \cdot d$
\State $\widetilde{x_{1}}$ = \text{\scriptsize EXPLORE\_NUM\_CELLS}($x_{1}$)
\If{$f(\widetilde{x_{1}}) > f(\widetilde{x_{0}}) \textbf{ and } x_{1} \in \Omega$}
        \Do
            \State j = j + 1
            \State $t_{j} = x_{j}$
            \State $x_{j + 1} = x_{j} + t_{j} \cdot d$
            \State $\widetilde{x_{j + 1}}$ = \text{\scriptsize EXPLORE\_NUM\_CELLS}($x_{j + 1}$)
        \doWhile{$f(\widetilde{x_{j + 1}}) > f(\widetilde{x_{j}}) + \epsilon \textbf{ and } x_{j + 1} \in \Omega$}
    \Else
        \State $x_{1} = x_{0}$
        \Do
            \State j = j + 1
            \State $t_{j} = -\frac{x_{j}}{2}$
            \State $x_{j + 1} = x_{j} + t_{j} \cdot d$
            \State $\widetilde{x_{j + 1}}$ = \text{\scriptsize EXPLORE\_NUM\_CELLS}($x_{j + 1}$)
        \doWhile{$f(\widetilde{x_{j + 1}}) \geq f(\widetilde{x_{j}}) \textbf{ and } x_{j + 1} \in \Omega$}
\EndIf
\State \textbf{Output: } $\widetilde{x_{j}}$
\end{algorithmic}
\end{algorithm}

\begin{figure}[!h]
    \centering
    \includegraphics[width=0.5\textwidth]{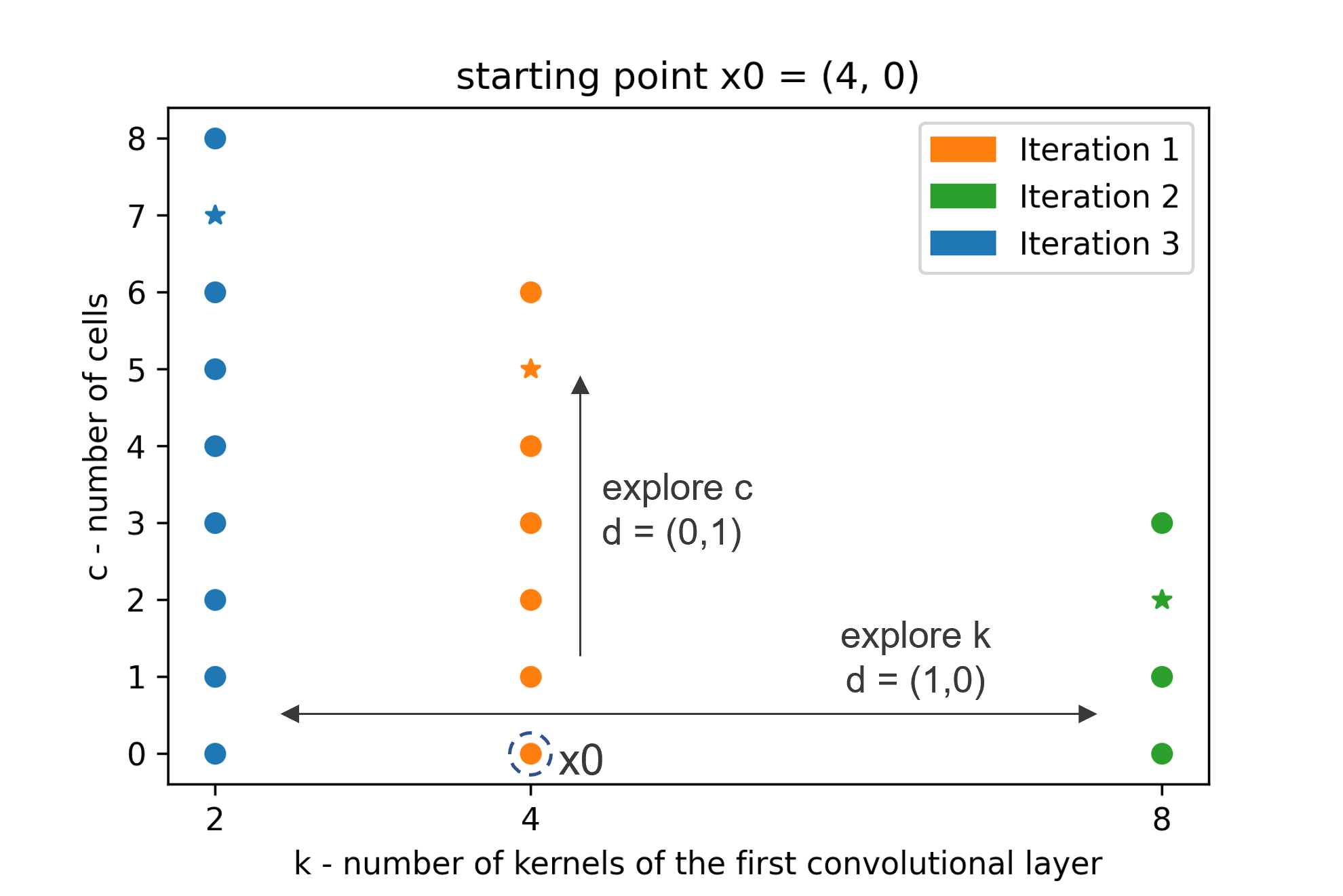}
    \caption{A figure that is showing the points explored by the algorithm during a sample run, in which the generalization capability does not increase during iteration 1 and gets worse during iteration 2. The star marker represents the network with the highest generalization capability for each iteration. As you can see, during each iteration the points move along the c axis. After each iteration, the starting point moves along the k axis.
    }
    \label{fig:plot}
\end{figure}

It is an alternating search in the direction of the two main axes, as shown in figure \ref{fig:plot}. First, it explores the axis of the cells' additions (direction $d = (0,1)$) given a starting point, using algorithm \ref{alg:explore}. Then it moves the starting point on the axis of the number of kernels of the first layer (direction $d = (1,0)$) and repeats the search. 

If the hardware selected for deployment allows, the number of kernels used in the first layer is doubled, i.e. the starting point $x_{0}$ is doubled. If the network found with the new starting point is better than the previous one, the algorithm continues doubling the starting point until the generalization capability improves, if the hardware limits are not reached. Otherwise, the algorithm continues by halving down the initial number of kernels, i.e. the starting point $x_{0}$, until performance degradation is met or an unfeasible point is found. $\widetilde{x_{j}}$ represents the algorithm's output.

\begin{algorithm}[!h]
\caption{\text{\scriptsize EXPLORE\_NUM\_CELLS}($x$)}\label{alg:explore}
\begin{algorithmic}
    \Require $x \in \Omega$
    \Ensure $\begin{cases} 
   f(x_{i}) > f(x_{i - 1}), x_{i} \in \Omega & \text{always} \\
   f(x_{i + 1}) \leq f(x_{i}) & \text{if } x_{i + 1} \in \Omega
\end{cases}$
    \State $x_{0} \gets x$, $d \gets (0, 1)$, $t \gets 1$, $i \gets 0$
    \Do
        \State $x_{i + 1} = x_{i} + t \cdot d$
        \State i = i + 1
    \doWhile{$f(x_{i + 1}) > f(x_{i}) \textbf{ and } x_{i + 1} \in \Omega$}
    \State \textbf{return} $x_{i}$
\end{algorithmic}
\end{algorithm}

Procedure \ref{alg:explore} explores the axis of the cells' additions ($d = (0,1)$) given a starting point $x$, with a unitary step size ($t = 1$). Practically, it continues to add cells until the generalization capability increases, if the new network is feasible. 

For the sake of synthesis, in algorithm \ref{alg:search}, x1 is explored in any case, even if it is not feasible. For the same reason the feasibility check for the point $x_{j + 1}$ is done after the exploration. Something similar happens in algorithm \ref{alg:explore}, where the network $x_{i + 1}$ is trained even if it is not feasible. 

\section{Experimental Methods}
\label{sec:methodology}
Five task-specific classification problems are chosen to evaluate the efficacy of the proposed technique. First the models obtained by using ColabNAS are compared with the ones obtained by applying TL, a common way to design lightweight task-specific CNN \cite{DeepTransferLearning}. The aim of this comparison is to show that ColabNAS is able to obtain much smaller models than TL at the cost of a few percentage points of accuracy on the test set, in an acceptable amount of time. Then, three different MCUs are used to show case the ability to adapting to different hardware, in low-RAM conditions, of ColabNAS. Finally, a comparison with state-of-the-art HW NAS, based on the Visual Wake Word dataset, a standard TinyML benchmark \cite{TinyML_benchmarks}, is presented.

Code, models and datasets are publicly available in the form of Google Colaboratory's notebooks at the link in section \ref{sec:introduction}. All the computations have been executed on Google Colaboratory using Tesla T4 GPUs, as of today, the ones mostly offered to free subscription users. 

The five classification problems have been chosen to provide a set of common task-specific datasets, publicly available with a variety of sizes spanning from a few thousand images to hundreds of thousand of images, also ensuring the possibility of reproducing the experiments proposed in this paper on free-subscription online GPU programs. Where test splits were not provided, they have been built using a 0.2 test split. i.e. 80\% of data for training, 20\% for testing. A brief description of the tasks follows hereafter.

\subsection{Melanoma Skin Cancer}
The Melanoma Skin Cancer classification task aims to discriminate between benign and malignant images of melanoma skin cancer. It is composed of a training set containing 9,605 images and a test set of 1,000 images. It is publicly available on kaggle under the name of ``Melanoma Skin Cancer Dataset of 10000 Images'' \footnote{\url{https://www.kaggle.com/datasets/hasnainjaved/melanoma-skin-cancer-dataset-of-10000-images?resource=download}}. 

\subsection{Visual Wake Words}
The Visual Wake Words classification task by \citet{vww} aims to discriminate between images with and without human presence. It is composed of a training set containing 115,000 and a test set containing 8,000 images, derived by the ``minival ids'' as done by \citet{vww}.

\subsection{Animals-3}
The Animals-3 classification task aims to discriminate between three species of animals: horse, butterfly and hen. It contains 2,623 instances of horses, 2,112 instances of butterflies, and 3,098 instances of hens. It is a subset of the Animals-10 dataset, which is publicly available on kaggle\footnote{\url{https://www.kaggle.com/datasets/alessiocorrado99/animals10}}.

\subsection{Flowers-4}
The Flowers-4 classification task aims to discriminate between four classes of flowers: dandelion, iris, tulip, and magnolia. It contains 1,052 instances of dandelion, 1,054 instances of iris, 1,048 instances of tulip, and 1,048 instances of magnolia. It is a subset of the Flowers dataset, which is publicly available on Kaggle \footnote{\url{https://www.kaggle.com/datasets/l3llff/flowers}}.

\subsection{MNIST}
The MNIST \cite{MNIST} classification task aims to discriminate between ten handwritten digits: 0, 1, 2, \dots, 9. It is a subset of a larger set available from NIST. It is composed of a training set containing 60,000 images and of a test set containing 10,000 images.

\section{Comparison with the networks obtained using Transfer Learning} \label{sec:comparison_with_the_networks_obtained_using_transfer_learning}
This section compares ColabNAS with TL. To perform TL, MobileNetV2 \cite{Mobilenetv2}, with frozen weights trained on ImageNet1k \cite{ImageNet}, is used as the backbone. The extracted features are then compressed by a bi-dimensional global average pooling layer and then given to a shallow classifier, composed of only one dense layer. The pre-processing pipeline is the same of ColabNAS, reported in section \ref{sec:technique}. It first applies both horizontal flips and random rotations, then applies min-max standardization, and finally uses batch normalization. Such models are trained for 20 epochs using a learning rate of $10^{-3}$ and a batch size of $128$. Then, the trained model is fine-tuned for $10$ epochs, with the backbone unfrozen, using a learning rate of $10^{-5}$.

Instead, ColabNAS is run using as constraints the characteristics of the models obtained using TL, i.e. RAM occupancy, ROM occupancy, and MACC. Candidate solutions are trained for $100$ epochs with a learning rate of $10^{-3}$ and a batch size of $128$. The same input size of 224x224x3, imposed by pre-trained MobileNetV2, is used for all the datasets. Results for the Visual Wake Words and MNIST dataset are not included in this comparison because of their high number of images which makes ColabNAS exceed the maximum amount of time guaranteed for free users on Google Colaboratory. 

The metrics used for the comparison are the test accuracy, the peak RAM occupancy, the Flash occupancy and the number of MACC. Post training quantization (PTQ) has been applied to all the models. X-CUBE-AI software from ST Microelectronics has been used to measure the peak RAM occupancy and the Flash occupancy, while the number of MACC has been computed using Keras backend of Tensorflow.

\subsection{Melanoma Skin Cancer}
Table \ref{tab:comparison_tl_melanoma} shows the comparison for the Melanoma Skin Cancer classification task. In this case, ColabNAS outperformed the result obtained by applying TL, by all means. ColabNAS produced a network that is 3.1 per cent points more precise, occupies 4.6 times less RAM, 35.2 times less Flash and performs 5.2 times less MACC compared to the TL's result. 

\begin{table}[!h]
\centering
 \begin{tabular}{l | c c c c | c }
               & Test Acc. & RAM   & Flash & MACC & Search Cost \\ 
               & [\%]      & [kiB] & [kiB] & [MM] & [hh]:[mm]   \\ \hline
     TL        & 88        & 2,523 & 2,650 & 300  & 00:16       \\
     Our       & 91.1      & 547   & 75.2  & 44   & 03:52       \\ \hline
               &           &       &       &      &             
 \end{tabular}
 \caption{\label{tab:comparison_tl_melanoma} Test accuracy, peak RAM occupancy, Flash occupancy, MACC and search cost of the resulting models of TL and ColabNAS (our proposed method) for the Melanoma Skin Cancer dataset.}
\end{table}

\subsection{Animals-3}
Table \ref{tab:comparison_tl_animals} shows the comparison for the Animals-3 classification task. In this case, ColabNAS produced a model that occupies 2.6 times less RAM, 13.4 times less Flash and performs 2 times less MACC while being 6 per cent points less precise compared to the TL's result. 

\begin{table}[!h]
\centering
 \begin{tabular}{l | c c c c | c }
               & Test Acc. & RAM   & Flash & MACC & Search Cost \\ 
               & [\%]      & [kiB] & [kiB] & [MM] & [hh]:[mm]   \\ \hline
     TL        & 99.2      & 2,523 & 2,653 & 300  & 00:11       \\
     Our       & 93.2      & 988   & 197.3 & 153  & 03:02       \\ \hline
               &           &       &       &      &             
 \end{tabular}
 \caption{\label{tab:comparison_tl_animals} Test accuracy, peak RAM occupancy, Flash occupancy, MACC and search cost of the resulting models of TL and ColabNAS (our proposed method) for the Animals-3 dataset.}
\end{table}

\subsection{Flowers-4}
Finally, table \ref{tab:comparison_tl_flowers} shows the comparison for the Flowers-4 classification task. In this case, ColabNAS produced a model that occupies 4.6 times less RAM, 44.4 times less Flash and performs 6.8 times less MACC while being 5.4 per cent points less precise compared to the TL's result. 

\begin{table}[!h]
\centering
 \begin{tabular}{l | c c c c | c }
               & Test Acc. & RAM   & Flash & MACC & Search Cost \\ 
               & [\%]      & [kiB] & [kiB] & [MM] & [hh]:[mm]   \\ \hline
     TL        & 99.2      & 2,523 & 2,653 & 300  & 00:07       \\
     Our       & 93.8      & 546   & 59.7  & 44   & 01:20       \\ \hline
               &           &       &       &      &             
 \end{tabular}
 \caption{\label{tab:comparison_tl_flowers} Test accuracy, peak RAM occupancy, Flash occupancy, MACC and search cost of the resulting models of TL and ColabNAS (our proposed method) for the Flowers-4 dataset.}
\end{table}

\section{Evaluation of the hardware-aware feature}
\label{sec:evaluation_of_the_hardware-aware_feature}
This section evaluates the hardware-aware feature of ColabNAS by providing results for different hardware targets. To highlight the ColabNAS feature of providing lightweight CNNs, three low-RAM STMicroelectronics (STM) MCUs from the Ultra-low Power series have been selected: the L010RBT6 (abbr. L0), the L151UCY6DTR (abbr. L1) and the L412KBU3 (abbr. L4). The generated networks were evaluated using the STM32 X-Cube-AI software.

\begin{table}[!h]
        \begin{tabular}{l | c c c} 
                        STM32 MCU     & RAM [kiB] & Flash [kiB] & CoreMark  \tabularnewline \hline
                        L010RBT6      & 20        & 128         & 75        \tabularnewline 
                        L151UCY6DTR   & 32        & 256         & 93       \tabularnewline 
                        L412KBU3      & 40        & 128         & 273 
        \end{tabular} \\ 
        \caption{\label{tab:nucleo_boards} Available RAM and Flash, and the CoreMark score of each hardware target considered for the experiment.}
\end{table}

Table \ref{tab:nucleo_boards} summarizes the key features of each MCU, i.e., available RAM and Flash, and CoreMark score, a benchmark for comparing performances of commercial MCUs. The values in table \ref{tab:nucleo_boards} set the constraints for running ColabNAS on each target. RAM and Flash were provided as is, while the MACC upper bound was obtained multiplying by $10^{4}$ the CoreMark score of the target, in order to allow a fair exploration of the search space. 50x50x3 input size has been used to cope with MCUs constrained resources. The results for each dataset follow.

\subsection{Melanoma Skin Cancer}
Table \ref{tab:hardware_awareness_melanoma} shows the resulting models for each target for the Melanoma Skin Cancer dataset. As can be seen, ColabNAS was able to provide a feasible model for each target, adapting to the resource available. The larger the target's resources the larger the model is. Models show a mean test accuracy drop of 1.8 per cent points while lowering the target's resources. The drop is more significant while passing from L1, the medium target, to L0, the small target.  

\begin{table}[!h]
        \begin{tabular}{c | c c c c c} 
                        Target                   & Acc  & RAM   & Flash & MACC  & Search Cost  \tabularnewline 
                        abbr.                    & [\%] & [kiB] & [kiB] & [k]   & [hh]:[mm]    \tabularnewline \hline
                        L0                       & 86.5 & 19.5  & 8.3   & 92    & 00:17        \tabularnewline 
                        L1                       & 88.7 & 22    & 14.4  & 654   & 00:19        \tabularnewline 
                        L4                       & 90.1 & 32.5  & 31.84 & 2,075 & 00:17        \tabularnewline \hline
                                                 &      &       &       &       &             
        \end{tabular} \\ 
        \caption{\label{tab:hardware_awareness_melanoma} Test accuracy, RAM and Flash occupancy, MACC and search cost of the resulting models for each target, indicated in its abbreviated form on the first column, for the Melanoma Skin Cancer dataset.}
\end{table}

\subsection{Visual Wake Words}
Table \ref{tab:hardware_awareness_vww} shows the resulting models for each target for the Visual Wake Words dataset. As can be seen, ColabNAS was able to provide a feasible model for each target, adapting to the resource available. The larger the target's resources the larger the model is. Models show a mean test accuracy drop of 4.2 per cent points while lowering the target's resources. The drop is more significant while passing from L1, the medium target, to L0, the small target. 

\begin{table}[!h]
        \begin{tabular}{c | c c c c c} 
                        Target                   & Acc  & RAM   & Flash & MACC  & Search Cost  \tabularnewline 
                        abbr.                    & [\%] & [kiB] & [kiB] & [k]   & [hh]:[mm]    \tabularnewline \hline
                        L0                       & 69.4 & 19    & 8.02  & 227   & 2:11         \tabularnewline 
                        L1                       & 74.5 & 22.5  & 18.5  & 657   & 3:04         \tabularnewline 
                        L4                       & 77.8 & 33    & 44.9  & 2,086 & 2:47         \tabularnewline \hline
                                                 &      &       &       &       &             
        \end{tabular} \\ 
        \caption{\label{tab:hardware_awareness_vww} Test accuracy, RAM and Flash occupancy, MACC and search cost of the resulting models for each target, indicated in its abbreviated form on the first column, for the Visual Wake Words dataset.}
\end{table}

\subsection{Animals-3}
Table \ref{tab:hardware_awareness_animals} shows the resulting models for each target for the Animals-3 dataset. ColabNAS was able to provide a feasible model for each target, adapting to the resource available. The larger the target's resources the larger the model is. Models show a mean test accuracy drop of 8.65 per cent points while lowering the target's resources. The drop is more significant while passing from L1, the medium target, to L0, the small target.  

\begin{table}[!h]
        \begin{tabular}{c | c c c c c} 
                        Target                   & Acc  & RAM   & Flash & MACC  & Search Cost  \tabularnewline 
                        abbr.                    & [\%] & [kiB] & [kiB] & [k]   & [hh]:[mm]    \tabularnewline \hline
                        L0                       & 67.9 & 19    & 8.03  & 227   & 00:09        \tabularnewline 
                        L1                       & 75.8 & 22.5  & 18.65 & 657   & 00:17        \tabularnewline 
                        L4                       & 85.2 & 33    & 44.86 & 2,086 & 00:09        \tabularnewline \hline
                                                 &      &       &       &       &             
        \end{tabular} \\ 
        \caption{\label{tab:hardware_awareness_animals} Test accuracy, RAM and Flash occupancy, MACC and search cost of the resulting models for each target, indicated in its abbreviated form on the first column, for the Animals-3 dataset.}
\end{table}

\subsection{Flowers-4}
Table \ref{tab:hardware_awareness_flowers} shows the resulting models for each target for the Flowers-4 dataset. As can be seen, ColabNAS was able to provide a feasible model for each target, adapting to the resource available. The larger the target's resources, the larger the model is. Models show a mean test accuracy drop of 6.05 per cent points while lowering the target's resources. In this case, the drop is more significant while passing from L4, the large target, to L1, the medium target.  

\begin{table}[!h]
        \begin{tabular}{c | c c c c c} 
                        Target                   & Acc  & RAM   & Flash & MACC  & Search Cost  \tabularnewline 
                        abbr.                    & [\%] & [kiB] & [kiB] & [k]   & [hh]:[mm]    \tabularnewline \hline
                        L0                       & 79.2 & 18.5  & 6.67  & 211   & 00:05        \tabularnewline 
                        L1                       & 84.4 & 21.5  & 10.91 & 633   & 00:08        \tabularnewline 
                        L4                       & 91.3 & 32.5  & 31.91 & 2,075 & 00:10        \tabularnewline \hline
                                                 &      &       &       &       &             
        \end{tabular} \\ 
        \caption{\label{tab:hardware_awareness_flowers} Test accuracy, RAM and Flash occupancy, MACC and search cost of the resulting models for each target, indicated in its abbreviated form on the first column, for the Flowers-4 dataset.}
\end{table}

\subsection{MNIST}
Table \ref{tab:hardware_awareness_MNIST} shows the resulting models for each target for the MNIST dataset. ColabNAS was able to provide a feasible model for each target, adapting to the resource available. The larger the target's resources the larger the model is. Models show a mean test accuracy drop of 4.9 per cent points while lowering the target's resources. The drop is more significant while passing from L1, the medium target, to L0, the small target.  

\begin{table}[!h]
        \begin{tabular}{c | c c c c c} 
                        Target                   & Acc  & RAM   & Flash & MACC  & Search Cost  \tabularnewline 
                        abbr.                    & [\%] & [kiB] & [kiB] & [k]   & [hh]:[mm]    \tabularnewline \hline
                        L0                       & 88.2 & 19.5  & 9.79  & 233   & 01:07        \tabularnewline 
                        L1                       & 95.6 & 22.5  & 18.80 & 657   & 01:41        \tabularnewline 
                        L4                       & 98   & 33    & 45.23 & 2,087 & 01:29        \tabularnewline \hline
                                                 &      &       &       &       &             
        \end{tabular} \\ 
        \caption{\label{tab:hardware_awareness_MNIST} Test accuracy, RAM and Flash occupancy, MACC and search cost of the resulting models for each target, indicated in its abbreviated form on the first column, for the MNIST dataset.}
\end{table}

\subsection{Final Considerations}
ColabNAS proven to be able to adapt to targets with low RAM availability, providing feasible solutions tailored to the available resources. The obtained models seem to depend more on the target's characteristics than on the dataset. Chosen a target, resulting model's RAM and MACC seem to remain almost constant across the different datasets explored in this section, while the flash occupancy shows higher variations. The search time appears proportional to the number of images in the dataset, given the fixed input size. By further lowering the input size, it should be possible to obtain networks with even smaller RAM occupancies, probably at the cost of further reducing test accuracies.

\section{Comparison with state-of-the-art hardware-aware NAS techniques on the Visual Wake Word dataset} \label{sec:comparison_with_state_of_the_art_hardware_aware_NAS_on_the_Visual_Wake_Word_dataset}
This section compares ColabNAS with two state-of-the-art HW NAS techniques for CNNs: MicroNets by \citet{micronets} and MCUNet by \citet{mcunet}. The comparison is based on the Visual Wake Word dataset \cite{vww}, a standard benchmark for TinyML models \cite{TinyML_benchmarks}. Imagenet1k \cite{ImageNet} is not included, given its general-purpose nature, which is not compatible with the task-specific nature of our method. 

Only the model with lowest RAM occupation has been selected from both projects\footnote{In the case of MicroNets the ``vww2\_50\_50\_INT8.tflite'' model has been downloaded from the ARM's GitHub web page \url{https://github.com/ARM-software/ML-zoo/blob/master/models/visual_wake_words/micronet_vww2/tflite_int8/vww2_50_50_INT8.tflite}. In the case of MCUNet, the ``mcunet-10fps-vww'' model has been downloaded from the laboratory's web page \url{https://hanlab.mit.edu/projects/tinyml/mcunet/release/mcunet-10fps_vww.tflite}}. Then, we ColabNAS has been ran on the Visual Wake Word dataset using the same hardware target of Micronets, the one having the smallest RAM among the targets of the two projects: the STMF446RE MCU having 128 kiB of RAM, 512 kiB of Flash and a CoreMark score of 608. The input size is also the same as Micronets: 50x50x3.

\begin{table}[!h]
        \begin{tabular}{c | c c c c c} 
                        \multirow{2}{*}{Project} & Acc  & RAM   & Flash  & Latency & Input                  \tabularnewline 
                                                 & [\%] & [kiB] & [kiB]  & [mS]    & Size                   \tabularnewline \hline
                        MCUNet                   & 87.4 & 168.5 & 530.52 & 2.16    & 64x64x3                \tabularnewline 
                        Micronets                & 76.8 & 70.5  & 273.81 & 1.15    & 50x50x3                \tabularnewline 
                        Our                      & 77.6 & 31.5  & 20.83  & 0.432   & 50x50x3                \tabularnewline \hline
                                                 &      &       &        &       &             
        \end{tabular} \\ 
        \caption{\label{tab:comparison_state_of_the_art} Test accuracy, RAM and Flash occupancy, latency and input size for the chosen models from Micronets, MCUNET and ColabNAS (our) for the Visual Wake Words dataset.}
\end{table}

\begin{figure}
    \centering
    \includegraphics[width=0.48\textwidth]{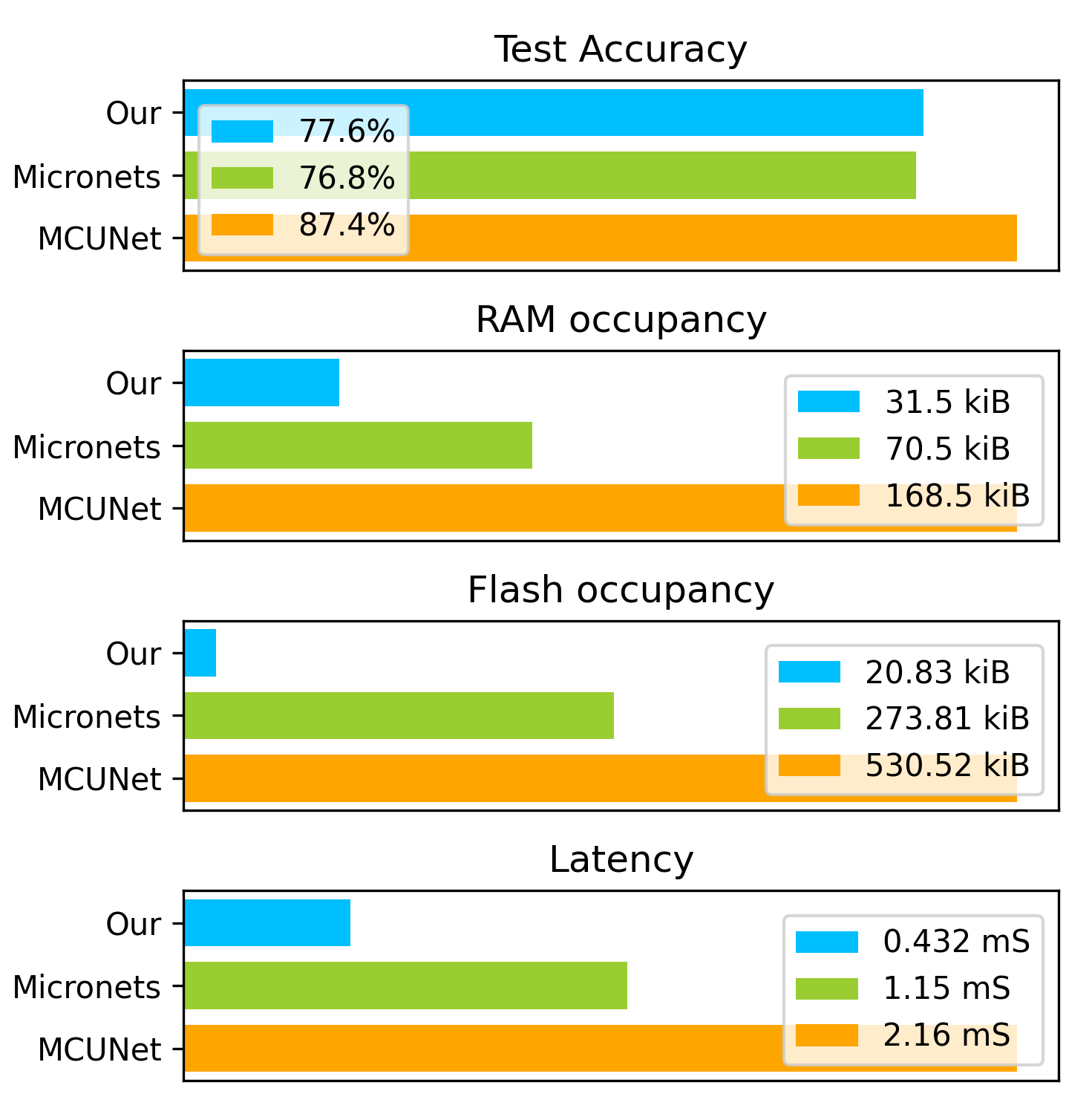}
    \caption{Graphical representation of test accuracy, RAM and Flash occupancy, latency and input size for the chosen models from Micronets, MCUNET and ColabNAS (our) for the Visual Wake Words dataset.}
    \label{fig:sota}
\end{figure}

Table \ref{tab:comparison_state_of_the_art} shows the comparison results, which are also graphically represented in figure \ref{fig:sota}. All the models are in TFLite format. They are fully quantized to perform 8-bit inference. Test accuracy, RAM occupancy and Flash occupancy are compared, as in the previous cases. Since there is no direct way to measure MACC using the TFLite API, no value for them is presented. To have a comparison of the execution times, a latency value is measured using the IPython magic command ``\%timeit'' alongside the TFLite interpreter invocation. The input tensor content is random. All the measurements were performed during the same Google Colaboratory session with a dual-core Intel(R) Xeon(R) CPU at 2.20GHz. No GPU was involved. 

Our model improves by all means the one proposed by Micronets. However, the solution proposed by MCUNet is still more accurate. It offers 9.8 per cent points accuracy higher than our solution while occupying 5.35 times more RAM, 25.47 times more Flash and being 5 times slower. %It offers 9.8 per cent points accuracy higher than our solution, the latter occupying 5.35 times less RAM, 25.47 times less Flash and being 5 times faster.

\subsection{Search Costs Comparison}

As stated by \citet{mcunet} MCUNet spent 300 GPU hours to produce its architecture. Instead, MicroNets by \citet{micronets} does not declare the time spent on finding the architecture. However, they run DNAS for 200 epochs to find it. DNAS by \citet{DARTS}, uses 1.5 GPU days (does not include the selection cost (1 GPU day) or the final evaluation cost by training the selected architecture from scratch (1.5 GPU days)) to find a network over the CIFAR-10 dataset, which contains 60000 32x32 colour images in 10 classes, with 6000 images per class. In this case, DNAS is run for 100 epochs. Considering that the visual wake word training set is composed of 107,954 images, that the input volume used is 50x50x1 and that the epochs used are double, the time spent on it is higher than the time spent on the CIFAR-10. Hence ColabNAS results considerably faster than MCUNet and MicroNets.

\section{Conclusion}
\label{sec:conclusion}
In this paper, we proposed ColabNAS: an affordable HW NAS technique for designing task-specific CNNs, able to work with low-RAM MCUs. Its novel search strategy, inspired by Occam's razor, has a low search cost. It obtains state-of-the-art results on the visual wake word dataset in just 3.1 GPU hours, improving, by all means, the solution found by Micronets.

The low search cost allows its execution on free online services such as Kaggle Kernel or Google Colaboratory, without owning a high-end GPU. Given the current global chip shortage, this can help those end-users and researchers who want to approach application of lightweight CNNs for custom classification problems.

It also provides task-specific lightweight CNNs that occupy on average 3.9 times less RAM, 31 times less Flash, and have 4.6 less MAC than those obtained by applying TL and FT, based on MobileNetV2 on the same task, one of the most used techniques to obtain task-specific lightweight CNNs in research. This is at the cost of losing 2.8 per cent accuracy points on average. 
Such a trade-off can target deployment on the growing field of wearable and distributed devices with embedded electronics hardware. Moreover, reducing the computational cost of the final application turns into more efficient power consumption. In a world where artificial intelligence is becoming more and more diffused, reducing energy consumption is a fundamental step towards a more sustainable future.

\end{document}